\documentclass{article}

\usepackage{arxiv}
\usepackage{natbib}
\usepackage[T1]{fontenc}    
\usepackage{hyperref}       
\usepackage{url}            
\usepackage{booktabs}       
\usepackage{amsfonts}       
\usepackage{nicefrac}       
\usepackage{microtype}      
\usepackage{lipsum}		
\usepackage{graphicx}
\usepackage{doi}
\usepackage{amsmath}
\usepackage[caption=false,font=normalsize,labelfont=sf,textfont=sf]{subfig}
\usepackage{multirow}

\title{Efficient Knowledge Deletion from Trained Models through Layer-wise Partial Machine Unlearning}

\author{ \href{https://orcid.org/0000-0003-2171-9623}{\includegraphics[scale=0.06]{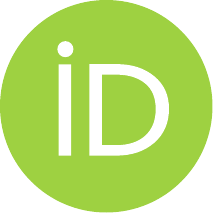}\hspace{1mm}Vinay Chakravarthi Gogineni}
\\
	Applied AI and Data Science\\
	The Mærsk Mc-Kinney Møller Institute\\
        University of Southern Denmark\\
	Camusvej 55, Odense M, 5230 \\
	\texttt{vigo@mmmi.sdu.dk} \\
	\And
	\href{https://orcid.org/0000-0003-2613-2696}{\includegraphics[scale=0.06]{orcid.pdf}\hspace{1mm}Esmaeil Nadimi} \\
	Applied AI and Data Science\\
	The Mærsk Mc-Kinney Møller Institute\\
        University of Southern Denmark\\
	Camusvej 55, Odense M, 5230 \\
	\texttt{esi@mmmi.sdu.dk} \\
}

\renewcommand{\shorttitle}{\textit{arXiv} Template}

\hypersetup{
pdftitle={A template for the arxiv style},
pdfsubject={q-bio.NC, q-bio.QM},
pdfauthor={David S.~Hippocampus, Elias D.~Striatum},
pdfkeywords={First keyword, Second keyword, More},
}

\begin{document}
\maketitle

\begin{abstract}
Machine unlearning has garnered significant attention due to its ability to selectively erase knowledge obtained from specific training data samples in an already trained machine learning model. This capability enables data holders to adhere strictly to data protection regulations. However, existing unlearning techniques face practical constraints, often causing performance degradation, demanding brief fine-tuning post unlearning, and requiring significant storage. In response, this paper introduces a novel class of machine unlearning algorithms. First method is \emph{partial amnesiac unlearning}, integration of layer-wise pruning with amnesiac unlearning. In this method, updates made to the model during training are pruned and stored, subsequently used to forget specific data from trained model. The second method assimilates \emph{layer-wise partial-updates} into label-flipping and optimization-based unlearning to mitigate the adverse effects of data deletion on model efficacy. Through a detailed experimental evaluation, we showcase the effectiveness of proposed unlearning methods. Experimental results highlight that the partial amnesiac unlearning not only preserves model efficacy but also eliminates the necessity for brief post fine-tuning, unlike conventional amnesiac unlearning. Moreover, employing layer-wise partial updates in label-flipping and optimization-based unlearning techniques demonstrates superiority in preserving model efficacy compared to their naive counterparts.
\end{abstract}


\section{Introduction}
In the current data-centric era, data holders actively gather valuable information from various sources, encompassing personal information of individuals such as images, speech, text and medical records. Machine learning (ML) \cite{ML1, DL1, DL2} has orchestrated a profound transformation in navigating and analyzing the immense reservoir of data available to data holders. ML facilitates the extraction of valuable insights from these extensive and information-rich repositories, thereby enhancing user experiences and providing an overall improved service in numerous applications. Furthermore, in practical, real-world scenarios, data pipelines often demonstrate dynamic behavior, consistently acquiring fresh data. Life-long ML leverages this continuous inflow of information to dynamically update knowledge and fine-tune ML model parameters \cite{LifelongML}. 

On the contrary, there are situations where the imperative for data deletion becomes obvious. Envision a scenario where a patient, who initially disclosed a specific medical condition with the healthcare system, later decides to retract this particular information due to privacy reasons. Similarly, users, attentive to privacy, may wish to erase their purchase history from an e-commerce platform, especially if they made unusual or atypical purchases during a specific period. Provisions like the \emph{right to be forgotten} in general data protection regulation (GDPR) \cite{GDPR}, empowers individuals to exert control over their personal data and imposes a legal obligation for the removal of such data when requested. 

In situations where data has already been used to train an ML model, the mere act of deleting the data from data holder's storage is inadequate. This inadequacy stems from the inherent ability of ML models, particularly complex deep neural networks (DNNs), to potentially memorize intricate patterns present within the training data \cite{DLmemory1, DLmemory2}. These models can retain information in their parameters, inadvertently revealing information about the data they were trained on. This vulnerability is exploited through various attacks like the membership inference attack \cite{MembershipIA1, MembershipIA2} and the model inversion attacks \cite{ModelIA1, ModelIA2}. Therefore, it is essential for a trained ML model to go through an unlearning process, purging the knowledge acquired from the data targeted for unlearning, often referred to as \emph{targeted data}. This necessity prompts an exploration into the realm of {\sl machine unlearning}; a process that selectively removes the influence of specific training data samples from an already trained ML model \cite{mul1, mul2}. The objective of this process is to ensure that, after unlearning, the model demonstrates behavior akin to a model that has never been trained on the target data.

Machine unlearning proves invaluable not only in safeguarding privacy but also instrumental in enhancing model security and resilience against adversarial attacks, especially data-poisoning attacks \cite{dpa1}. In such attacks, adversaries inject carefully crafted malicious data into the training set to influence the model's behavior and undermine its trustworthiness. Machine unlearning serves as a defense mechanism against these attacks by selectively forgetting manipulated data, thus restoring the model's integrity. Additionally, machine unlearning enhances the adaptability of models in dynamic environments over time. It becomes indispensable in selectively discarding outdated knowledge acquired from obsolete data, enabling models to stay agile and relevant in evolving scenarios. This strategic forgetting of obsolete information is paramount, ensuring that the model remains robust and efficient.

Numerous studies in the literature have investigated the nascent field of machine unlearning. These efforts can be broadly categorized into two groups: {\sl exact unlearning} and {\sl approximate unlearning} methods \cite{VerifiableMUL}. Exact unlearning methods completely erase the influence of a target data on existing ML model. This is achieved by excluding the target data samples from the training data set and then retraining the model. As unlearning requests persist, the need to retrain the model from scratch becomes resource-intensive, particularly due to the training of large-scale DNNs on data reservoirs. To tackle this challenge, several efficient procedures for swift retraining \cite{SQL, DaRE, DeltaGrad, SISA, Descent-to-Delete} have been proposed. On the other hand, approximate unlearning methods entail modifying the existing model parameters in a way that the model behaves as if it has never seen the target data. Selective forgetting through scrubbing \cite{scrubbing}, linear filtration \cite{linear-filtration}, certified removal \cite{certified-removal}, and amnesiac unlearning \cite{amnesiacML} are a few examples of this kind. A subgroup of approximate unlearning methods perceives unlearning as an optimization problem \cite{OptimMUL1, OptimMUL2, FastYE, ZeroShotMU},
modifying the existing model parameters by maximizing the loss related to the target data samples. Nevertheless, the majority of these unlearning works are compromised by one or more of the ensuing issues:
\begin{enumerate}
    \item The majority of existing approximate unlearning techniques e.g.,  \cite{amnesiacML, scrubbing, FastYE}, exhibit a  degradation in model performance after handling an unlearning request. This degradation becomes more pronounced with an increase in the number of target data samples. 
    \item A few other existing unlearning techniques demand substantial storage capacity to facilitate the deletion of target data samples. This issue arises from the retention of storage-intensive models trained on data subsets \cite{SISA, Descent-to-Delete} or the storage of updates made to model parameters during training \cite{amnesiacML}.
    \item Most prevailing unlearning techniques make use of \emph{retained data}, i.e., the original training data excluding the targeted data, during the unlearning process. Certain unlearning techniques, exemplified in \cite{amnesiacML, FastYE}, involve a brief fine-tuning phase on the retained data for a limited number of iterations to restore model performance as it was before the unlearning.  
\end{enumerate}

\subsection{Our Contributions}
This paper aims to address the aforementioned concerns while effectively erasing the impact of targeted data from an already trained model. Our contributions are as follows:
\begin{itemize}
    \item We advocate for adopting layer-wise pruning in tandem with conventional amnesiac unlearning to efficiently erase the targeted data from an already trained model. The proposed method, termed \emph{partial amnesiac unlearning}, selectively discards a portion of the updates made to the model layer-by-layer basis during the training phase and stores them for subsequent use in the unlearning phase. In the unlearning phase, subtracting the layer-wise pruned updates from the trained model efficiently erases the impact of target data while preserving the model's efficacy on retained data. A critical aspect of the proposed partial amnesiac unlearning is that it does not require any brief fine-tuning following the unlearning phase. Furthermore, storing the pruned updates significantly reduces storage space requirements in contrast to conventional amnesic unlearning \cite{amnesiacML}.
    \item We suggest integrating \emph{layer-wise partial-updates}  \cite{partialLMS} into label-flipping- and optimization-based unlearning. Employing layer-wise partial updates during the unlearning phase strategically mitigates the potential negative impact on model efficacy. This approach aims to maintain the model's effectiveness on retained data while efficiently erasing the impact of target data from the trained model.
    \item We conduct a comprehensive empirical assessment to demonstrate the effectiveness of the proposed unlearning methods, wherein the membership inference metric serves as a key performance indicator. Our analysis encompasses diverse data sets such as MNIST \cite{mnist} and MEDMNIST \cite{medmnist}. Additionally, we explore various neural network architectures, including multilayer perceptron (MLP) with $2$ fully connected layers, a basic random convolutional neural network (ConvNet) with $2$ convolutional layers and $3$ fully connected layers, the well-known Lenet \cite{lenet}, AlexNet \cite{alexnet}, and $9$-layer residual network (ResNet$9$) \cite{resnet}. Our experimental results emphasize the superiority of the proposed methods over their conventional counterparts   
\end{itemize}

\section{Related Work}
A naive method to ensure information erasure involves discarding target data samples from the training data set and then retraining an ML model from scratch. However, initiating retraining from scratch for each unlearning request would demand a substantial amount of resources and time. To devise a more efficient alternative, various efforts have been made in the literature. The concept of unlearning was initially introduced in \cite{SQL} to facilitate the forgetting of training data samples from trained ML model. This approach represents the learning algorithm in a summation form, where each summation is the sum of transformed data samples. To fulfill an unlearning request, this approach simply updates the summations, making it faster than retraining from scratch. In \cite{DaRE}, a variant of random forests, referred to as data removal-enabled (DaRE) forests, was proposed. DaRE forests allow the removal of training data with minimal retraining. The DeltaGrad algorithm, presented in \cite{DeltaGrad}, facilitates the swift retraining of stochastic gradient-based machine learning algorithms when forgetting a limited set of data samples. DeltaGrad is applicable for strong convex and smooth objective functions. In \cite{SISA}, a deterministic deletion procedure named sharded, isolated, sliced, and aggregated training (SISA) has been introduced. In the SISA framework, the entire data set undergoes random partitioning into $K$ non-overlapping parts, commonly referred to as shards, which are further divided into slices. Each shard independently undergoes model training, and the resulting separate models are subsequently averaged to produce the final model. Notably, SISA mitigates the necessity for complete retraining, as the model can be specifically retrained on the shard where the deletion request has been initiated. However, SISA is susceptible to membership inference attacks, particularly when an adversary possesses the ability to observe both the model before and after the removal of a specific user's data point \cite{Chen}. This concern is addressed in \cite{Descent-to-Delete} by implementing a differential privacy mechanism over the shard models before aggregating them to generate the final model. All these retraining techniques are termed as exact unlearning methods, as the resultant model has not been exposed to the target data.

Another category of unlearning algorithms, known as approximate unlearning methods, modifies the already trained model parameters to ensure that the model behaves as if it has never encountered the target data. Approximate unlearning methods circumvent the need for retraining, requiring fewer resources compared to exact unlearning methods. In \cite{scrubbing}, a quadratic scrubbing procedure based on Newton's update has been proposed. This method uses the identities of gradient and Hessian to selectively erase information from an already trained model. The linear filtration method proposed in \cite{linear-filtration} applies a linear transformation to the logits of a classifier, designed to handle class-wide deletion requests in a computationally efficient manner. In \cite{certified-removal}, a certified data removal mechanism has been proposed based on Newton's update. This method ensures the removal of data up to the level of differential privacy guarantees. The amnesiac unlearning method proposed in \cite{amnesiacML} stores the details of data samples presented in each batch, along with their corresponding contributions to the model during training. In the unlearning phase, the contributions made by the target data samples are subtracted from the trained model. In \cite{OptimMUL1, OptimMUL2}, the unlearning process involves retraining the already trained model for a limited number of steps on the target data. These strategic approaches aim to maximize the loss instead of minimizing it. Motivated by the error-minimizing noise framework \cite{mnoise}, a method for generating error-maximization noise in the context of unlearning has been introduced in \cite{FastYE}. This method learns the noise that maximizes the loss on target data samples. In \cite{ZeroShotMU}, error-minimization and maximization concepts are integrated for efficient unlearning. A novel adaptive machine unlearning approach leveraging attention mechanisms has been introduced in \cite{attentionul}. This approach computes attention scores for each data sample in the data set and incorporates these scores into the unlearning process. The exploration of unlearning concepts has been extended into the realm of graph data in \cite{gul}. A preliminary extension of machine unlearning within federated network settings has been explored in \cite{fedul1, fedul2}.

\section{Preliminaries}
In this section, we provide a concise overview of machine learning, delve into the intriguing concept of machine unlearning, and present a very brief overview of selective prominent machine unlearning algorithms relevant to our work. We systematically introduce the essential notation, paving the foundation for a clear presentation of the proposed algorithms. It is worth to emphasize that our focus throughout this paper is specifically on supervised learning.
\subsection{Machine Learning}
Consider a training data set $\mathcal{D}$ consists of $N$ number of data pairs $\{{\bf x}_i, y_i\}_{i=1}^{N}$; where ${\bf x}_i$ is $i$th input data sample (e.g., an image) and $y_i$ is its corresponding reference (e.g., class label). The objective in supervised learning is to identify the nonlinear function $f$, parametrized by a DNN model ${\bf w}$, which characterizes the relationship between input data samples ${\bf x}_i$ and their corresponding references $y_i$. This relationship is expressed mathematically as $f: \mathcal{X} \to \mathcal{Y}$, where $\mathcal{X}$ and $\mathcal{Y}$ represent the spaces encompassing all input data samples and references, respectively.

Learning the DNN model parameters ${\bf w}$ entails minimizing the empirical risk $\mathcal{L}(\mathcal{D}, {\bf w})$, measure how effectively the model ${\bf w}$ maps data samples ${\bf x}_i$ to their corresponding references ${\bf y}_i$, for $i = 1, 2, \ldots, N$. For example, in the context of classification task, cross-entropy is a widely recognized loss function, defined as:
\begin{equation}
\mathcal{L}(\mathcal{D}, {\bf w}) = - \frac{1}{N} \sum_{i=1}^{N} \sum_{c=1}^{C} y_{i, c} \cdot \log(f({\bf x}_i)_c) ,  
\end{equation}
where $C$ denotes the number of classes and $y_{i, c} =1$ if the true label for data sample ${\bf x}_i$, i.e., $y_i$ is class $c$ and $0$ otherwise. The predicted probability of the DNN model for class $c$ given the input ${\bf x}_i$ is denoted as $f({\bf x}_i)_c$. The model parameters ${\bf w}$ can be learned recursively by following a steepest descent search on $\mathcal{L}(\mathcal{D}, {\bf w})$. The batch gradient descent recursion for updating the model parameters is: 
\begin{equation}
 {\bf w}_e =  {\bf w}_{e-1} - \alpha \hspace{1mm} \nabla_{{\bf w}} \mathcal{L}(\mathcal{D}, {\bf w})\Big|_{{\bf w}_{e-1}}, 
\end{equation}
where ${\bf w}_e$ and ${\bf w}_{e-1}$ are the model parameters from the current and previous epochs, respectively. $\nabla_{{\bf w}} \mathcal{L}(\mathcal{D}, {\bf w})$ is the gradient of the loss function with respect to model parameters ${\bf w}$ and $\alpha$ is the learning rate. To improve computational efficiency, it is often preferable to use minibatches of data $\mathcal{D}^{b}$, for $b=1, 2, \ldots, B$, rather than the entire data set $\mathcal{D}$ during training. The minibatch gradient descent recursion for updating the model parameters during epoch $e$ is:
\begin{equation}
 {\bf w}_{e, b} =  {\bf w}_{e, b-1} - \alpha \hspace{1mm} \nabla_{{\bf w}} \mathcal{L}(\mathcal{D}^{b}, {\bf w})\Big|_{{\bf w}_{e, b-1}}, 
\end{equation}
with ${\bf w}_{e, 0} = {\bf w}_{e-1}$. After processing the final minibatch with index $B$, the updated model parameters ${\bf w}_{e, B}$ initiate the subsequent epoch $e+1$. This process continues until convergence is achieved.
\subsection{Machine Unlearning}
Consider a scenario where a DNN model, parameterized by ${\bf w}$, is trained on a data set $\mathcal{D}$. Let $\mathcal{D}_t$ be the targeted data set, which includes only the data samples designated for deletion. Additionally, let $\mathcal{D}_r$ is the retained data set, which is the original training data set with excluding targeted data, i.e., $\mathcal{D}_r = \mathcal{D} \setminus \mathcal{D}_t$, where $\setminus$ is the set minus operator. The goal of machine unlearning is to erase the impact of the target data $\mathcal{D}_t$ from the trained model ${\bf w}$. Subsequently, the resulting model post-unlearning, denoted as ${\bf w}^{\prime}$, is expected to perform on the retained data set $\mathcal{D}_r$ similar to the model before unlearning.

\subsubsection{Sharded, Isolated, Sliced, Aggregation (SISA):}
The SISA training exhibits some resemblance to that of distributed training strategies. In SISA training, the original data set $\mathcal{D}$ is partitioned into $S$ disjoint shards denoted as $\mathcal{D}_{s}$, for $s=1, 2, \ldots, S$. Each shard $\mathcal{D}_{s}$, for $s=1, 2, \ldots, S$, is subsequently divided into $R$ slices denoted as $\mathcal{D}_{s, r}$, for $r=1, 2, \ldots, R$. The training process for each shard $s$ begins with the initial training of a model ${\bf w}_{s, 1}$ on the first slice $\mathcal{D}_{s, 1}$, with the resulting parameters are stored. Subsequently, the model, initially trained on the first slice ${\bf w}_{s, 1}$, undergoes additional training on the second slice $\mathcal{D}_{s, 2}$, and the parameters of the resulting model ${\bf w}_{s, 2}$ are stored. This iterative process continues until training is completed on the $R$th slice and the parameters of ${\bf w}_{s, R}$ are stored. The final model, denoted as ${\bf w}_{s, R}$, represents the shard model ${\bf w}_{s}$. Finally, the shard models ${\bf w}_{s}$, for $s=1, 2, \ldots, S$, are aggregated to form a comprehensive final model. 

When an unlearning request arises to delete specific data samples, retraining becomes necessary only for the shard model whose shard contains the target data samples. The retraining process can be swiftly performed by utilizing the model parameters, stored before the training on the slice containing the data sample marked for unlearning.

SISA exhibits several limitations. Firstly, if the data set is partitioned in a non-i.i.d manner, the resulting final model accuracy tends to be notably low. Secondly, when unlearning requests arise simultaneously on multiple shards, the retraining process demands a substantial amount of computational resources and time. Lastly, the performance of the final model is significantly dependent on the number of shards and slices employed.  
\subsubsection{Amnesiac Unlearning:}
Amnesiac unlearning selectively erases acquired knowledge associated with the targeted data. During the training process, the data holder, responsible for training an ML model, actively keeps track of the details of data samples within the data set, precisely noting which data samples appear in each batch. Additionally, the data holder also stores the corresponding updates made to the model parameters. This systematic storage of information serves as a crucial repository, ensuring a comprehensive record of the model's learning journey.

Upon receiving an unlearning request, the data holder takes action by removing the updates incorporated into the model from the specific batches containing the targeted data samples. Let ${\bf w}$ be an already trained model, and $\mathcal{B}_t$ denotes the set of batches containing the targeted data samples. Then, the amnesic unlearning generate the new model ${\bf w}^{\prime}$ by effectively removing the impact of the targeted data from the already trained model ${\bf w}$ as \cite{amnesiacML}:
\begin{equation}
    {\bf w}^{\prime}_{} =  {\bf w} - \sum\limits_{e=1}^{E}\sum\limits_{b \in \mathcal{B}_t }^{} \Delta {\bf w}_{e,b}  ,
\end{equation}
where $E$ denotes the number of epochs that the training phase has been carried out and $\Delta {\bf w}_{e, b}$ represents the contribution made by minibatch $b$ to the model parameters ${\bf w}$ during epoch $e$ of the training phase.

Amnesiac unlearning has proven to be effective in erasing the influence of a limited number of targeted data samples on trained model. As the number of affected batches increases, the resulting model efficacy after amnesiac unlearning diminishes proportionally. Furthermore, amnesiac unlearning exhibits diminished model efficacy on retained data when higher batch sizes are employed. It is important to emphasize that when a larger number of batches is affected, the model requires a brief retraining phase following the amnesic unlearning step to restore its performance. Notably, the implementation of amnesic learning demands substantial storage space to accommodate the tracking of updates made to the model parameters during training.

\subsubsection{Label-Flipping-based Unlearning:}
The objective of unlearning through label-flipping is to intentionally obscure the model's comprehension of the targeted data, ensuring it possesses no valuable knowledge about that specific data. In this method, incorrect labels are randomly assigned to the targeted data samples. Subsequently, the network undergoes retraining through multiple iterations using this modified targeted data to fortify the intentional obfuscation. To unlearn an entire class, every example within that class needs to be assigned with a randomly chosen incorrect label. On the other hand, to unlearning a particular set of examples, assigning a few examples with erroneous labels is sufficient. The study in \cite{amnesiacML} recommends to retrain the model on the modified data only for a very few iterations for efficient unlearning. 
\subsubsection{Optimization-based Unlearning:}
In the optimization-based unlearning, the network undergoes a few iterations of retraining on the targeted data set $\mathcal{D}_t$ with objective to maximize the empirical loss instead of minimizing it. The optimization-based unlearning obtains the modified model, denoted as ${\bf w}^{\prime}$, from the already trained model ${\bf w}$ using the batch gradient descent rule as follows:
\begin{equation}
 {\bf w}^{\prime}_e =  {\bf w}^{\prime}_{e-1} + \alpha^{\prime} \hspace{1mm} \nabla_{{\bf w}^{\prime}} \mathcal{L}(\mathcal{D}_{t}, {\bf w}^{\prime})\Big|_{{\bf w}^{\prime}_{e-1}}, 
\end{equation}
where ${\bf w}^{\prime}_{0} = {\bf w}$ and $\alpha^{\prime}$ is the learning rate of the retraining. 

Although label-flipping- and optimization-based unlearning methods efficiently erase the knowledge of targeted data within trained models, the resultant model exhibits poor performance on retained data compared to the pre-unlearning.

\section{Proposed Machine Unlearning Methods}
In this section, we present the proposed class of machine unlearning algorithms.
\subsection{Partial Amnesiac Unlearning}
Subtracting the entire contribution made by specific batches containing targeted data has a detrimental effect on the model's behavior. The observed adverse effect stems from the fact that the update from specific batches includes contributions not only from the targeted data but also from the retained data presented in those batches. Hence, the resultant model from conventional amnesiac unlearning exhibits poor performance on retained data, necessitating a brief fine-tuning to restore model performance. However, this fine-tuning is again resource-intensive, especially for large-scale DNNs. Furthermore, to keep track of the updates made to the model during training phase, conventional amnesiac unlearning requires large-amount of storage space. The proposed partial amnesiac unlearning effectively tackles these challenges by incorporating layer-wise pruning into the conventional amnesiac unlearning. The proposed partial amnesiac unlearning operates as follows. 

During training phase, partial amnesiac unlearning prunes the updates made to the model layer-by-layer basis and store the pruned updates instead of retaining the entire update. Let $\Delta{\bf w}_{e, b}$ represent the contribution made by batch $b$ during epoch $e$, and $\Delta {\bf w}_{l, e,b}$ denote the update corresponding to the $l$th layer in the DNN model. Then, partial amnesiac unlearning prunes $\Delta {\bf w}_{l, e,b}$ and obtains its pruned version $\widetilde{\Delta {\bf w}}_{l, e,b}$ as 
\begin{equation}
\widetilde{\Delta {\bf w}}_{l, e,b}= {\bf P}_l \odot \Delta{\bf w}_{l, e, b},  
\end{equation}
where ${\bf P}_l$ is the pruning matrix whose size is the same size as that of the $l$th layer of the DNN model. The elements of ${\bf P}_l$ are either $1$ or $0$. The symbol $\odot$ denotes the elementwise product or Hadamard product operator. The position of ones in ${\bf P}_l$ dictates which weights in the updates are going to be subtracted from the trained model. Furthermore, the number of zeros in ${\bf P}_l$ is determined by the pruning percentage. In a nutshell, partial amnesiac unlearning sets certain weights in the update for each layer to zero before storing it. This process repeats for every layer in the DNN model until the model ${\bf w}$ is trained. It is important to highlight that storing the pruned update requires less storage space compared to storing the entire update. As a result, partial amnesiac unlearning demands less storage compared to its conventional counterpart.

During the unlearning phase, partial amnesic unlearning subtracts the stored pruned updates $\widetilde{\Delta {\bf w}}_{e,b}$ corresponding to the specific batches containing the targeted data. Let $\mathcal{B}_t$ denotes the set of batches containing the data samples of targeted data. Then, the partial amnesic unlearning generate the new model ${\bf w}^{\prime}$ from the trained model ${\bf w}$ as:
\begin{equation}
    {\bf w}^{\prime}_{} =  {\bf w} - \sum\limits_{e=1}^{E}\sum\limits_{b \in \mathcal{B}_t }^{} \widetilde{\Delta {\bf w}}_{e,b} .
\end{equation}

By selectively removing only a portion of the contribution from specific batches made to the model during the training phase, the proposed partial amnesiac unlearning minimizes the loss of knowledge acquired from the retained data in those batches. This approach guarantees the preservation of valuable knowledge from the retained data while effectively erasing the influence of targeted data from the model.

Within the framework of partial amnesiac unlearning, we explore three distinct pruning mechanisms, such as layer-wise random pruning, layer-wise magnitude-based pruning, and global pruning. Layer-wise random pruning involves randomly removing the parameters from the model. Layer-wise magnitude-based pruning, on the other hand, identifies and removes parameters based on the magnitude, typically removing those with lower magnitudes. Global pruning, in contrast, involves the removal of a fixed percentage of the least important parameters across the entire DNN model. It is important to highlight that when the pruning percentage for each layer of the DNN is set to zero, the proposed partial amnesiac unlearning transforms into conventional amnesiac unlearning.

Given that the initial layers of a DNNs play a crucial role in learning fundamental representations, a prudent approach is to employ more aggressive pruning in these initial layers of the update. As we traverse deeper into the network, the pruning percentage can be gradually decreased. This strategy aims to preserve the low-level representations of retained data within the model, preventing the inadvertent removal of essential learned features during the unlearning process. 

\subsection{Layer-wise Partial Updates induced Label-Flipping-based Unlearning}
Assigning incorrect labels to the targeted data set $\mathcal{D}_t$ and updating the trained model ${\bf w}$ for a limited number of iterations can effectively eliminate the influence of the targeted data from the trained model. However, it comes at the cost of altering the representations of the retained data. This is because, during the unlearning process, the model learns incorrect representations for the retained classes. To minimize the disturbance in learned representations of retained classes, it is beneficial to utilize layer-wise partial-update mechanism in the unlearning phase. 

Let $\mathcal{D}^{\prime}_{t}$ be the modified targeted data with incorrect labels. Layer-wise partial-updates induced label-flipping-based unlearning obtains the new model ${\bf w}^{\prime}$ from the trained model ${\bf w}$ as: 
\begin{equation}
 {\bf w}^{\prime}_e =  {\bf w}^{\prime}_{e-1} - \alpha^{\prime} \hspace{1mm} \widetilde{\nabla_{{\bf w}^{\prime}}} \mathcal{L}(\mathcal{D}^{\prime}_{t}, {\bf w}^{\prime})\Big|_{{\bf w}^{\prime}_{e-1}}, 
\end{equation}
where $\widetilde{\nabla_{{\bf w}^{\prime}}} \mathcal{L}(\mathcal{D}^{\prime}_{t}, {\bf w}^{\prime})$ represents the layer-wise partial gradient of the loss, i.e., a fraction of the actual gradient $\nabla_{{\bf w}^{\prime}} \mathcal{L}(\mathcal{D}^{\prime}_{t}, {\bf w}^{\prime})$ with ${\bf w}^{\prime}_{0} = {\bf w}$ (the original model parameters), and $\alpha^{\prime}$ is the learning rate of the unlearning. In $\widetilde{\nabla_{{\bf w}^{\prime}}} \mathcal{L}(\mathcal{D}^{\prime}_{t}, {\bf w}^{\prime})$, the $l$th layer partial gradient, denoted by $\widetilde{\nabla{{\bf w}^{\prime}_l}} \mathcal{L}(\mathcal{D}^{\prime}_{t}, {\bf w}^{\prime})$ is:
\begin{equation}\label{eq11}
\widetilde{\nabla{{\bf w}^{\prime}_l}} \mathcal{L}(\mathcal{D}^{\prime}_{t}, {\bf w}^{\prime}) = {\bf S}_{l, e} \odot \nabla_{{\bf w}^{\prime}_l} \mathcal{L}(\mathcal{D}^{\prime}_{t}, {\bf w}^{\prime}),    
\end{equation}
where $\nabla_{{\bf w}^{\prime}_l} \mathcal{L}(\mathcal{D}^{\prime}_{t}, {\bf w}^{\prime})$ is the actual gradient corresponds to $l$th layer of DNN model. ${\bf S}_{l, e}$ is the selection matrix, containing elements of either zero or one. The positions of ones in ${\bf S}_{l, e}$ determine which weights of the $l$th layer receive updates during epoch $e$. The selection matrix for the next epoch ${\bf S}_{l, e+1}$, can be obtained by performing a simple circular shift operation on the selection matrix of the current epoch ${\bf S}_{l, e}$.

In DNNs, the initial layers are responsible for learning fundamental representations. These representations, which are shared among various classes, significantly contribute to the model's ability to generalize. On the other hand, deeper layers learn more class-specific representations. In the case of label-flipping-based unlearning, where incorrect labels are assigned to the targeted class, it is beneficial to prioritize the update of weights in the initial layers. This emphasis ensures the preservation of fundamental features. Conversely, updating too many weights in the deeper layers may distort class-specific representations. Hence, when employing layer-wise partial updates into label-flipping-based unlearning method, it is advisable to configure the selection matrices to update a substantial number of parameters in the initial layers, gradually decreasing the number of updated parameters as the network depth increases. In this way it is possible to preserve essential generic features while mitigating the risk of distorting class-specific representations.

\subsection{Layer-wise Partial Updates induced Optimization-based Unlearning}
Updating the trained model ${\bf w}$ on the targeted data set $\mathcal{D}_t$ with the aim of maximizing the empirical loss effectively erases target data knowledge from trained model. However, this process corrupts the fundamental features shared among different classes. Utilizing layer-wise partial updates can mitigate the adverse effects of optimization-based unlearning on these fundamental features.

Layer-wise partial updates induced optimization-based unlearning obtains the new model ${\bf w}^{\prime}$ from the trained model ${\bf w}$ utilizing , as:
\begin{equation}
 {\bf w}^{\prime}_e =  {\bf w}^{\prime}_{e-1} + \alpha^{\prime} \hspace{1mm} \widetilde{\nabla_{{\bf w}^{\prime}}} \mathcal{L}(\mathcal{D}_t, {\bf w}^{\prime})\Big|_{{\bf w}^{\prime}_{e-1}}, 
\end{equation}
where $\widetilde{\nabla_{{\bf w}^{\prime}}} \mathcal{L}(\mathcal{D}_{t}, {\bf w}^{\prime})$ represents the layer-wise partial gradient of the loss, i.e., a fraction of the actual gradient $\nabla_{{\bf w}^{\prime}} \mathcal{L}(\mathcal{D}_{t}, {\bf w}^{\prime})$ with ${\bf w}^{\prime}_{0} = {\bf w}$ (the original trained model parameters), and $\alpha^{\prime}$ is the learning rate of the unlearning. In $\widetilde{\nabla_{{\bf w}^{\prime}}} \mathcal{L}(\mathcal{D}_{t}, {\bf w}^{\prime})$, the $l$th layer partial gradient, denoted by $\widetilde{\nabla{{\bf w}^{\prime}_l}} \mathcal{L}(\mathcal{D}_{t}, {\bf w}^{\prime})$ is same as given in \eqref{eq11}.

When employing layer-wise partial updates into optimization-based unlearning method, it is recommended to configure the selection matrices to update a less number of parameters in the initial layers, gradually increasing the number of updated parameters as the network depth increases. In this way it is possible to preserve essential generic features while selectively modifying class-specific representations, allowing the erasure of targeted class knowledge. 
\section{Experimental Results}
Extensive experiments were conducted to evaluate the efficacy of the proposed unlearning methods and provided comparison to their naive counterparts. We considered the model accuracy on the targeted data and retained data, known as \emph{membership inference metric} as performance indicator.  

\subsection{Experimental Hardware Setup} All experiments were performed in Python $3.11$ and use the PyTorch deep learning library \cite{pytorch}. The system was equipped with Apple M$2$ Pro chip comprising $12$-core CPU, $16$-core GPU and $16$-core neural engine with $16$GB RAM.
\subsection{Data Sets}
Two diverse data sets were considered in our experiments, one from computer vision domain and the other from medical domain. This deliberate choice allows for a thorough assessment of the effectiveness and generalizability of our proposed methods across varied domains, ensuring the reliability and applicability of our experimental results.
\begin{enumerate}
    \item MNIST Handwritten Digits data set \cite{mnist}: is a widely recognized benchmark datset, comprises a total of $70,000$ grayscale images, with $60,000$ images designated for training and $10,000$ images for testing. Each image is of size $28 \times28$ pixels depicting handwritten digits from $0$ to $9$. Each image is accompanied by corresponding labels indicating the digit represented. 
    \item MEDMNIST-OrganAMNIST Medical Image data set \cite{medmnist}: is a publicly available benchmark data set for medical image analysis. Derived from $3$D-computed tomography (CT) images sourced from the Liver Tumor Segmentation Benchmark (LiTS) \cite{LiTS}, the data set is tailored for multi-class classification of $11$ body organs. With a total of $58,850$ grayscale images, the data set is partitioned into $34,581$ training images, $6,491$ validation images, and $17,778$ testing images. Each image, depicting a body organ, possesses a resolution of $1\times 224\times 224$ pixels. The OrganAMNIST data set enhances the experimental evaluation of image classification algorithms, bringing richness through its diverse and comprehensive collection of medical domain images.
\end{enumerate}
\subsection{Network Architectures}
\begin{itemize}
    \item Multi-layer Perceptron (MLP): comprises two fully connected layers. The hidden layer consists of $500$ neurons.
    \item ConvNet: consists of three convolutional layers, each followed by a max pooling layer, and $3$ fully connected layers.
    \item LeNet: one of the foundational models in the development of modern CNNs \cite{lenet}. LeNet typically consists of seven layers, including three convolutional layers followed by two pooling layers and two fully connected layers. 
    \item AlexNet: winner of the ImageNet large scale visual recognition challenge (ILSVRC) in 2012, which significantly advanced the field of computer vision. Layer wise details of AlexNet can be found in \cite{alexnet}.
    \item ResNet$9$: is a variant of the residual network (ResNet) architecture, typically consists of several residual blocks, each containing convolutional layers and shortcut connections followed by a single fully connected layer \cite{resnet}.
    
\end{itemize}
\subsection{Assessment of Partial Amnesiac Unlearning}
\subsubsection{Model Behavior During Training and Unlearning Phases:} 
\begin{figure}[t!]
  \centering
  \includegraphics[width=\linewidth, height =65mm]{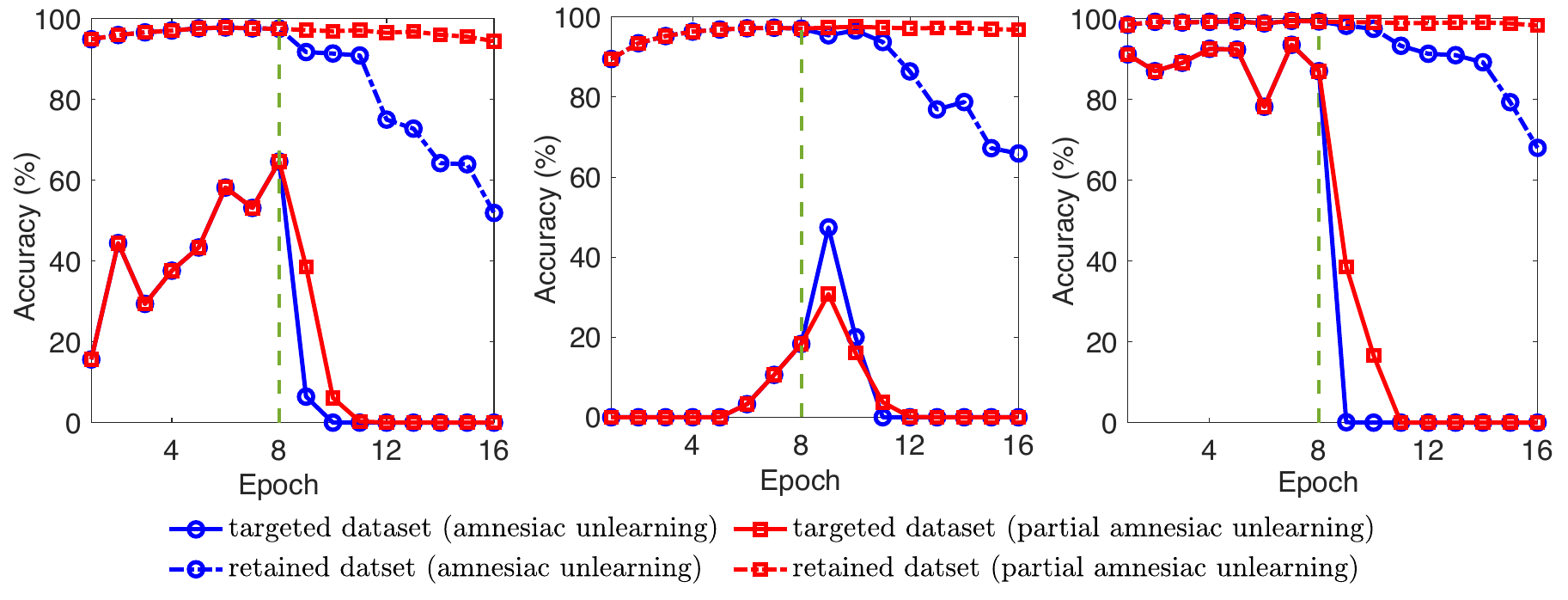}
  \caption{Model behavior during training and unlearning phases. Comparison of model accuracy between partial amnesiac unlearning and conventional amnesiac unlearning on targeted and retained MNIST data for three DNN architectures: MLP (left), LeNet (middle), and ResNet$9$ (right). The vertical dashed line indicates the point of the unlearning request initiation.}  \label{fig1}
\end{figure}

We investigated the effectiveness of proposed partial amnesiac unlearning in the context of member inference metric. To do so, we designated the data belonging to class-$3$ in MNIST data set as the targeted data, while the remaining data was considered to be the retained data. Initially, $2$-layer MLP, LeNet, and ResNet$9$ classifier models were trained on the MNIST data set for $8$ epochs utilizing a batch size of $128$ and a learning rate of $0.001$. These models validation accuracy on targeted and retained classes is illustrated in Figure \ref{fig1}. After $8$ epochs, an unlearning request was initiated to erase the knowledge of targeted class from trained classifier models. We simulated proposed partial amnesiac unlearning and its conventional counterpart, amnesiac unlearning, to process the unlearning request. It is worth mentioning that the conventional amnesiac learning was simulated without a brief fine-tune step after unlearning to ensure a fair comparison with the proposed method. The resulting model accuracy on targeted and retained classes is illustrated in Figure \ref{fig1}. From Figure \ref{fig1}, it is evident that the conventional amnesiac unlearning method successfully erase the impact of targeted data from trained models within $2$ to $3$ epochs; however, the resultant model exhibits performance drop on the retained data set. On the contrary, the proposed partial amnesiac unlearning method takes a few more epochs to completely remove the impact of targeted data. Nonetheless, the resultant model maintains a comparable accuracy on the retained data set, similar to its performance before the unlearning phase.

\begin{figure}[t!]
  \centering
  \includegraphics[width=\linewidth, height =65mm]{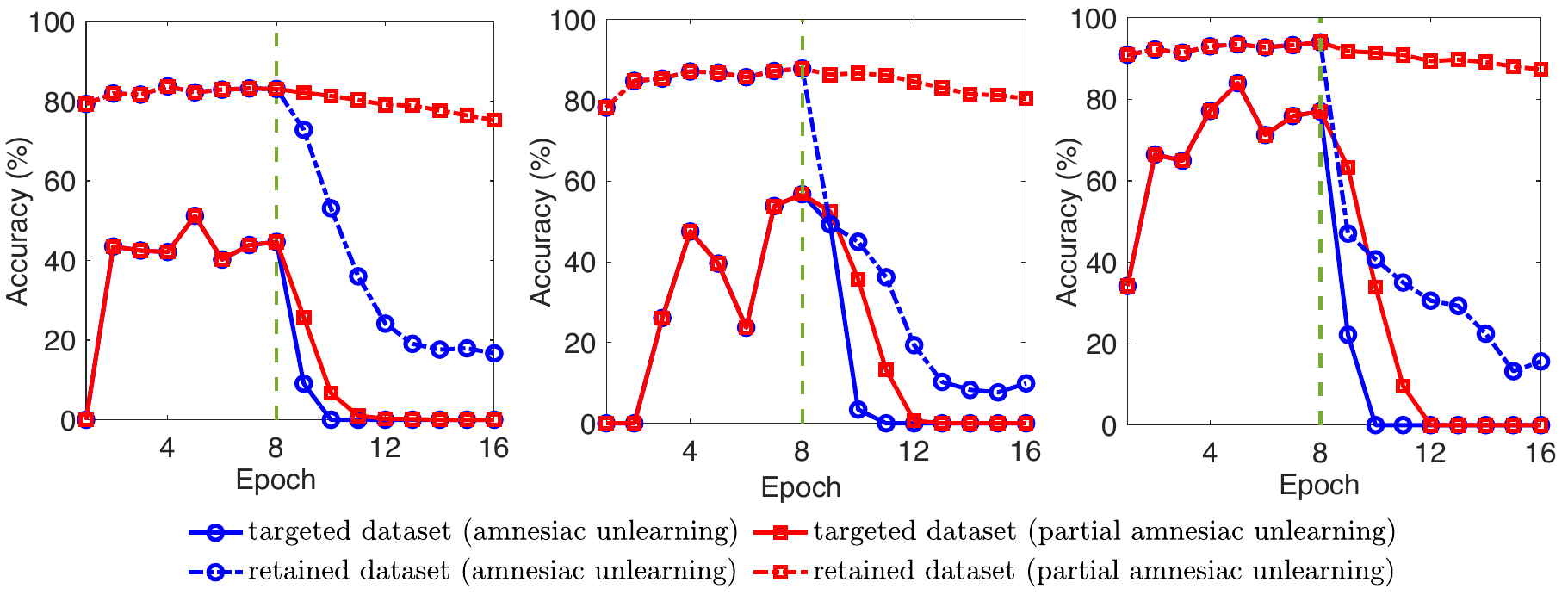}
  \caption{Model behavior during training and unlearning phases. Comparison of model accuracy between partial amnesiac unlearning and conventional amnesiac unlearning on targeted and retained OrganAMNIST data for three DNN architectures: ConvNet (left), AlexNet (middle), and ResNet$9$ (right). Vertical dashed line represents the unlearning request initiation.}
  \label{fig2}
\end{figure}

To further assess the performance of the proposed partial amnesiac unlearning, we conducted a similar experiment using the OrganAMNIST medical data set. After training the ConvNet, AlexNet, and ResNet$9$ classifier models for $8$ epochs, an unlearning request was triggered to erase the impact of class-$3$ data from trained models. Simulations of partial amnesiac unlearning and its conventional counterpart were conducted to process this unlearning request. The resulting model accuracy on both the targeted and retained data is illustrated in Figure \ref{fig2}. From Figure \ref{fig2}, we see that similar to the MNIST data set experiment, the conventional amnesiac unlearning method successfully removes the impact of the targeted data from the trained model within $2$ to $3$ epochs. However, the resulting model exhibits a significant performance drop on the retained data. In contrast, the proposed partial amnesiac unlearning method takes a few more epochs to erase the impact of target data. However, it maintains comparable accuracy on the retained data set, akin to its performance before the unlearning process. 

It is worth emphasizing that the proposed partial amnesiac unlearning achieves superior performance compared to its conventional counterpart without necessitating any fine-tuning after the unlearning process.

\subsubsection{Model Accuracy on Targeted and Retained Classes:} To elucidate the reduction in model effectiveness on the retained data after conventional amnesiac unlearning, we presented the model accuracy for certain retained classes (e.g., $2$, $7$, and $9$) in Table \ref{tabel1}. The results clearly indicates that the conventional amnesiac unlearning is successful in erasing the impact of targeted data from trained models, but adversely impacts the classification accuracy of retained classes. For example, in the case of the MLP, when erasing the knowledge of the targeted class (i.e., class-$3$) from the trained model, conventional amnesiac unlearning completely eradicated the ability to identify class-$7$. Similarly, in the case of LeNet and ResNet$9$, the model's capability to correctly identify class-$2$ is significantly compromised after the unlearning process. On the other hand, employing the proposed partial amnesiac unlearning for erasing targeted class from trained models shows a very minimal impact on the accuracy of the retained classes.

\begin{table}[t!]
\centering
\begin{tabular}{|c|c|c|c|ccc|}
\hline
\multirow{2}{*}{Model} & \multirow{2}{*}{Method} & \multirow{2}{*}{$\mathcal{D}_t$} & \multirow{2}{*}{$\mathcal{D}_r$} & \multicolumn{3}{c|}{Retained Classes} \\ \cline{5-7} 
                         &          &   &       & \multicolumn{1}{c|}{Class 2} & \multicolumn{1}{c|}{Class 7} & Class 9 \\ \hline
\multirow{2}{*}{MLP}     &  \cite{amnesiacML}        & 0 & 51.93 & \multicolumn{1}{c|}{80.91}   & \multicolumn{1}{c|}{0}       & 99.70   \\ \cline{2-7} 
                         & Proposed & 0 & 94.47 & \multicolumn{1}{c|}{97.96}   & \multicolumn{1}{c|}{84.04}   & 99.40   \\ \hline
\multirow{2}{*}{LeNet}   &  \cite{amnesiacML}        & 0 & 65.90 & \multicolumn{1}{c|}{54.45}   & \multicolumn{1}{c|}{96.01}   & 98.81   \\ \cline{2-7} 
                         & Proposed & 0 & 96.86 & \multicolumn{1}{c|}{96.51}   & \multicolumn{1}{c|}{95.52}   & 97.42   \\ \hline
\multirow{2}{*}{ResNet9} & \cite{amnesiacML}         & 0 & 68.04 & \multicolumn{1}{c|}{30.71}   & \multicolumn{1}{c|}{52.14}   & 95.83   \\ \cline{2-7} 
                         & Proposed & 0 & 98.33 & \multicolumn{1}{c|}{97.18}   & \multicolumn{1}{c|}{96.59}   & 99.40   \\ \hline
\end{tabular}
\caption{Comparison of model efficacy between partial amnesiac unlearning and conventional amnesiac unlearning on targeted and retained classes of MNIST data set.}\label{tabel1}
\vspace{-5mm}
\end{table}

\begin{table}[t!]
\centering
\begin{tabular}{|c|c|c|c|ccc|}
\hline
\multirow{2}{*}{Model} & \multirow{2}{*}{Method} & \multirow{2}{*}{$\mathcal{D}_t$} & \multirow{2}{*}{$\mathcal{D}_r$} & \multicolumn{3}{c|}{Retained Classes} \\ \cline{5-7} 
                         &          &   &       & \multicolumn{1}{c|}{Class 1} & \multicolumn{1}{c|}{Class 6} & Class 11 \\ \hline
\multirow{2}{*}{ConvNet} & \cite{amnesiacML}         & 0 & 16.66 & \multicolumn{1}{c|}{0}       & \multicolumn{1}{c|}{14.55}   & 0        \\ \cline{2-7} 
                         & Proposed & 0 & 75.20 & \multicolumn{1}{c|}{55.30}   & \multicolumn{1}{c|}{66.97}   & 58.91    \\ \hline
\multirow{2}{*}{AlexNet} & \cite{amnesiacML}         & 0 & 9.90  & \multicolumn{1}{c|}{0}       & \multicolumn{1}{c|}{46.71}   & 2.81     \\ \cline{2-7} 
                         & Proposed & 0 & 80.44 & \multicolumn{1}{c|}{61.96}   & \multicolumn{1}{c|}{74.30}   & 77.38    \\ \hline
\multirow{2}{*}{ResNet9} & \cite{amnesiacML}         & 0 & 15.64 & \multicolumn{1}{c|}{0}       & \multicolumn{1}{c|}{0}       & 69.85    \\ \cline{2-7} 
                         & Proposed & 0 & 87.34 & \multicolumn{1}{c|}{83.78}   & \multicolumn{1}{c|}{83.40}   & 99.04    \\ \hline
\end{tabular}
\caption{Comparison of model efficacy betwen partial amnesiac unlearning and conventional amnesiac unlearning on targeted and retained classes of OrganAMNIST data set.}\label{table2}
\vspace{-5mm}
\end{table}

The detrimental effects of conventional amnesiac unlearning are more highlighted in the case of OrganAMNIST data set. Table \ref{table2} displays the model accuracy for specific retained classes, such as $1$, $6$, and $11$. The findings in Table \ref{table2} underscore the significant adverse impact of conventional amnesiac unlearning on retained classes, with notably poor model efficacy for the majority of retained classes. In contrast, the proposed partial amnesiac unlearning method successfully erases the impact of the targeted class from the already  trained models with minimal negative influence on the accuracy of retained classes.

The rationale behind this lies in subtracting the pruned updates, as opposed to the entire update made during training, which minimizes the adverse effects on the model's behavior in partial amnesiac unlearning. Additionally, applying more aggressive pruning to the initial layers, with the pruning percentage decreasing as the network depth increases, shows no detrimental impact on the model's ability to learn fundamental features. Consequently, the model after partial amnesiac unlearning exhibits better performance on the retained data set and requires no brief fine-tuning, unlike in the case of conventional amnesiac unlearning.  

\subsubsection{Class Activation Maps:} 
\begin{figure}[t!]
\centering
\subfloat{\includegraphics[width=60mm, height =40mm]{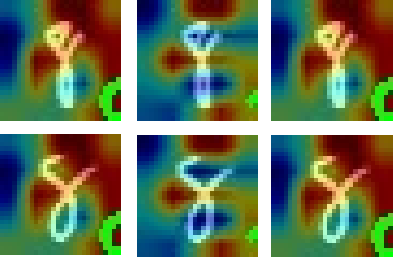}}\hspace{8mm}
\subfloat{\includegraphics[width=60mm, height =40mm]{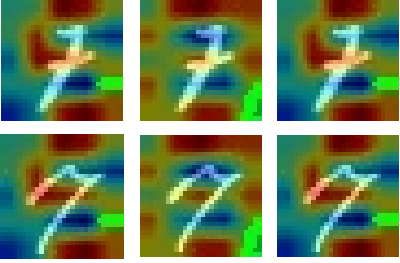}}
\caption{Class activation maps of LeNet model. Targeted classes are class-$3$ on the left side and class-$1$ on the right side. In each case, the first, second, and third  column images are the class activation maps before unlearning, after conventional amnesiac unlearning, and proposed partial amnesiac unlearning, respectively. Blue colour indicates low activation region while red colur indicates high activation regions.}\label{CAMs}
\end{figure}

To further examine the performance decline associated with conventional amnesiac unlearning, we leveraged class activation maps (CAMs). CAMs provide insights into the image regions that contribute to the prediction of a particular class. CAMs of LeNet model for images belonging to class-$8$ and class-$1$ before unlearning and corresponding CAMs of the same images after conventional amnesiac unlearning and the proposed partial amnesiac unlearning are presented in Figure. \ref{CAMs}. In CAMS presented in Figure. \ref{CAMs}, blue colour signifies low activation regions, indicating minimal influence on the predicted class, while red denotes high activation regions, representing significant contributions to the classification decision. From Figure. \ref{CAMs}, it is evident that the proposed partial amnesiac unlearning preserves the regions identified to be important for predicting the specific class. As a result the model performs well on retained classes without necessitating for brief-fine tuning post-unlearning. On the other hand, conventional amnesiac unlearning alters a few important regions to insignificant regions, resulting in poor performance on the retained classes. 
 
\subsubsection{Model Accuracy against Number of Affected Batches:}
\begin{figure}[t!]
  \centering
  \includegraphics[width=\linewidth, height =110mm]{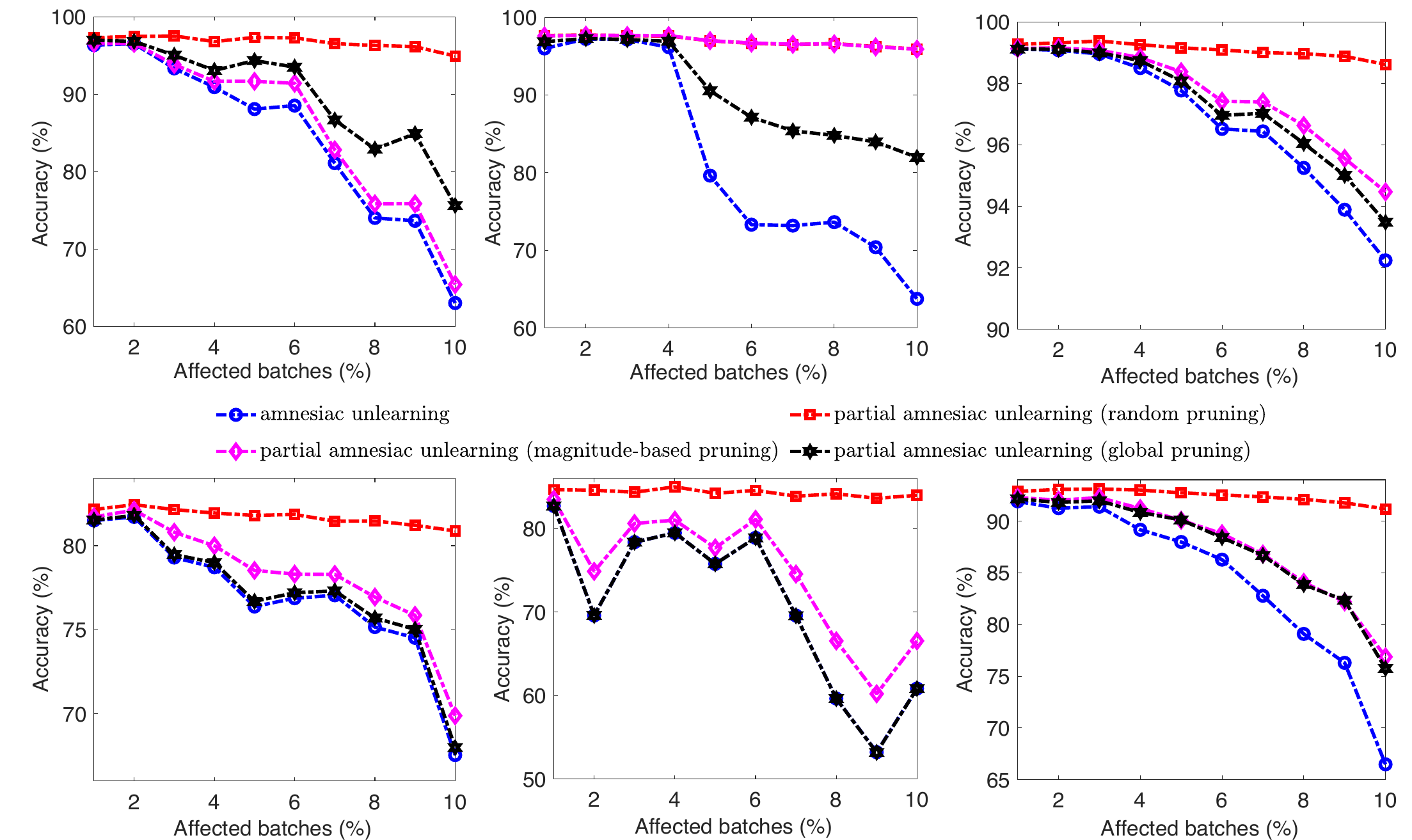}
  \caption{Comparison of model efficacy between partial amnesiac unlearning between conventional amnesiac unlearning against percentage of affected batches. }\label{fig3}
\end{figure}
Next, we investigated how the number of affected batches influences model efficacy in partial amnesiac learning. To do so, we plotted percentage of affected batches vs model accuracy on the retained data set for both the MNIST and OrganAMNIST in Figure \ref{fig3}. We presented the partial amnesiac unlearning curves with different pruning strategies. Figure \ref{fig3} clearly indicates that the deterioration of model efficacy on the retained data following conventional amnesiac unlearning becomes increasingly apparent as the number of affected batches rises. This decline in performance can be attributed to the intrinsic nature of conventional amnesiac unlearning, which not only removes the knowledge of targeted data from trained models but also eliminates knowledge acquired from the retained data. On the contrary, partial amnesiac unlearning adeptly mitigates this detrimental effect. This is accomplished by selectively removing the pruned updates, rather than discarding the entire updates made by the affected batches to the model during the training process. Subtracting pruned updates during training mitigates the loss of knowledge acquired from retained data in the affected batches. As a result, the partial amnesiac unlearning method showcases model efficacy comparable to the performance observed before the initiation of the unlearning process. Moreover, employing a random pruning strategy with the pruning amount decreasing along the network depth showcased superior performance compared to both magnitude-based and global pruning strategies.

\subsection{Evaluation of Layer-wise Partial-Updates induced Label-Flipping-based Unlearning}
To evaluate the effectiveness of layer-wise partial-updates induced label-flipping-based unlearning, we presented the model accuracy on the targeted class (i.e., class-$3$) and specific retained classes (such as $2$, $7$, and $9$) of MNIST data set in Table \ref{tabel3}. Additionally, we included the results of naive label-flipping-based unlearning for comparison. From Table \ref{tabel3}, it is evident that naive label-flipping-based unlearning effectively erases the knowledge of the targeted class from the trained model. However, it also has an adverse impact on a few retained classes, such as class $9$ in this case. 

\begin{table}[t!]
\centering
\begin{tabular}{|c|c|c|c|ccc|}
\hline
\multirow{2}{*}{Model} & \multirow{2}{*}{Method} & \multirow{2}{*}{$\mathcal{D}_t$} & \multirow{2}{*}{$\mathcal{D}_r$} & \multicolumn{3}{c|}{Retained Classes} \\ \cline{5-7} 
                         &          &      &       & \multicolumn{1}{c|}{Class 2} & \multicolumn{1}{c|}{Class 7} & Class 9 \\ \hline
\multirow{2}{*}{MLP}     & \cite{amnesiacML} & 0    & 90.75 & \multicolumn{1}{c|}{91.47}   & \multicolumn{1}{c|}{96.10}   & 71.55   \\ \cline{2-7} 
                         & Proposed         & 0.89 & 95.65 & \multicolumn{1}{c|}{94.86}   & \multicolumn{1}{c|}{96.20}   & 91.97   \\ \hline
\multirow{2}{*}{LeNet}   &   \cite{amnesiacML}       & 1.78 & 86.61 & \multicolumn{1}{c|}{95.93}   & \multicolumn{1}{c|}{83.17}   & 52.32   \\ \cline{2-7} 
                         & Proposed & 2.67 & 90.62 & \multicolumn{1}{c|}{97.38}   & \multicolumn{1}{c|}{82.39}   & 72.74   \\ \hline
\multirow{2}{*}{ResNet9} & \cite{amnesiacML}         & 0    & 97.28 & \multicolumn{1}{c|}{98.64}   & \multicolumn{1}{c|}{98.90}   & 84.14   \\ \cline{2-7} 
                         & Proposed & 0    & 97.88 & \multicolumn{1}{c|}{98.35}   & \multicolumn{1}{c|}{98.90}   & 90.58   \\ \hline
\end{tabular}
\caption{Comparison of model accuracy between layer-wise partial-updates induced label-flipping-based unlearning and its conventional counterpart on targeted and retained classes of MNIST data set.}\label{tabel3}
\vspace{-3mm}
\end{table}

\begin{table}[t!]
\centering
\begin{tabular}{|c|c|c|c|ccc|}
\hline
\multirow{2}{*}{Model} & \multirow{2}{*}{Method} & \multirow{2}{*}{$\mathcal{D}_t$} & \multirow{2}{*}{$\mathcal{D}_r$} & \multicolumn{3}{c|}{Retained Classes} \\ \cline{5-7} 
                         &         &      &       & \multicolumn{1}{c|}{Class 1} & \multicolumn{1}{c|}{Class 6} & Class 11 \\ \hline
\multirow{2}{*}{ConvNet} & \cite{amnesiacML} & 0    & 74.07 & \multicolumn{1}{c|}{69.30}   & \multicolumn{1}{c|}{44.58}   & 70.96    \\ \cline{2-7} 
                         &   Proposed       & 0    & 77.67 & \multicolumn{1}{c|}{72.20}   & \multicolumn{1}{c|}{53.33}   & 78.13    \\ \hline
\multirow{2}{*}{AlexNet} &   \cite{amnesiacML}       & 0    & 76.48 & \multicolumn{1}{c|}{77.22}   & \multicolumn{1}{c|}{56.64}   & 87.95    \\ \cline{2-7} 
                         & Proposed & 0    & 78.97 & \multicolumn{1}{c|}{79.92}   & \multicolumn{1}{c|}{56.33}   & 77.17    \\ \hline
\multirow{2}{*}{ResNet9} &   \cite{amnesiacML}       & 0    & 76.93 & \multicolumn{1}{c|}{97.87}   & \multicolumn{1}{c|}{78.32}   & 74.68    \\ \cline{2-7} 
                         & Proposed & 0.37 & 82.27 & \multicolumn{1}{c|}{97.20}   & \multicolumn{1}{c|}{73.68}   & 75.95    \\ \hline
\end{tabular}
\caption{Comparison of model accuracy between layer-wise partial-updates induced label-flipping-based unlearning and its conventional counterpart on targeted and retained classes of OrganAMNIST data set.}\label{tabel4}
\vspace{-3mm}
\end{table}

To delve deeper into this, we performed a similar experiment on the OrganAMNIST data set, and the corresponding model accuracy on the targeted class (i.e., class-$3$) and specific retained classes (such as $1$, $6$, and $11$) are presented in Table \ref{tabel4}. The results in Table \ref{tabel4} confirm the adverse effect of naive label-flipping-based unlearning on retained classes. In the case of OrganAMNIST, this adverse effect is more pronounced. However, leveraging layer-wise partial-updates in label-flipping-based unlearning mitigates this adverse effect and demonstrates superior performance compared to its naive counterpart, both in the MNIST and OrganAMNIST cases.

The reason for this is the naive label-flipping-based unlearning approach assigns random labels to target data in an attempt to modify the model. As a result, the resulting model not only learns incorrect features for the target class, which is beneficial for erasing the knowledge of the target class from the model, but also adversely affects the representation of retained classes. The utilization of layer-wise partial-updates mitigate this adverse effect by slowing down the process of modifying the representations of retained classes through partial updates.

\subsection{Evaluation of Layer-wise Partial-Updates in Optimization-based Unlearning}
To evaluate the impact of layer-wise partial-updates on optimization-based unlearning, we compared its model accuracy against its naive counterpart on targeted and specific retained classes of both MNIST and OrganAMNIST data sets. The corresponding results are presented in Tables \ref{tabel5} and \ref{tabel6}. The findings suggest that, although naive optimization-based unlearning effectively erases knowledge of the targeted data, it adversely affects the classification accuracy of certain retained classes. However, leveraging layer-wise partial-updates in optimization-based unlearning mitigates this adverse effect and demonstrates superior performance compared to its naive counterpart, both in the MNIST and OrganAmNIST cases.

\begin{table}[t!]
\centering
\begin{tabular}{|c|c|c|c|ccc|}
\hline
\multirow{2}{*}{Model} & \multirow{2}{*}{Method} & \multirow{2}{*}{$\mathcal{D}_t$} & \multirow{2}{*}{$\mathcal{D}_r$} & \multicolumn{3}{c|}{Retained Classes} \\ \cline{5-7} 
                         &          &      &       & \multicolumn{1}{c|}{Class 1} & \multicolumn{1}{c|}{Class 6} & Class 11 \\ \hline
\multirow{2}{*}{FCNN}    & \cite{OptimMUL2} & 0    & 85.12 & \multicolumn{1}{c|}{98.15}   & \multicolumn{1}{c|}{97.17}   & 40.13    \\ \cline{2-7} 
                         &  Proposed        & 0.39 & 91.36 & \multicolumn{1}{c|}{98.25}   & \multicolumn{1}{c|}{97.85}   & 63.42    \\ \hline
\multirow{2}{*}{LeNet}   &   \cite{OptimMUL2}       & 0.29 & 89.21 & \multicolumn{1}{c|}{93.60}   & \multicolumn{1}{c|}{81.61}   & 80.97    \\ \cline{2-7} 
                         & Proposed & 0.89 & 93.11 & \multicolumn{1}{c|}{95.15}   & \multicolumn{1}{c|}{87.06}   & 86.62    \\ \hline
\multirow{2}{*}{ResNet9} &  \cite{OptimMUL2}        & 0    & 97.33 & \multicolumn{1}{c|}{98.83}   & \multicolumn{1}{c|}{94.94}   & 96.03    \\ \cline{2-7} 
                         & Proposed & 0    & 97.93 & \multicolumn{1}{c|}{99.12}   & \multicolumn{1}{c|}{96.10}   & 97.22    \\ \hline
\end{tabular}
\caption{Comparison of model accuracy between layer-wise partial-updates induced optimization-based unlearning and its naive counterpart on targeted and retained classes of MNIST data set.}\label{tabel5}
\vspace{-3mm}
\end{table}

\begin{table}[t!]
\centering
\begin{tabular}{|c|c|c|c|ccc|}
\hline
\multirow{2}{*}{Model} & \multirow{2}{*}{Method} & \multirow{2}{*}{$\mathcal{D}_t$} & \multirow{2}{*}{$\mathcal{D}_r$} & \multicolumn{3}{c|}{Retained Classes} \\ \cline{5-7} 
                         &          &      &       & \multicolumn{1}{c|}{Class 1} & \multicolumn{1}{c|}{Class 6} & Class 11 \\ \hline
\multirow{2}{*}{ConvNet} & \cite{OptimMUL2} & 0.12 & 71.71 & \multicolumn{1}{c|}{92.56}   & \multicolumn{1}{c|}{50.73}   & 36.99    \\ \cline{2-7} 
                         & Proposed         & 0.50 & 74.61 & \multicolumn{1}{c|}{91.31}   & \multicolumn{1}{c|}{54.60}   & 43.09    \\ \hline
\multirow{2}{*}{AlexNet} &   \cite{OptimMUL2}       & 0.5  & 83.76 & \multicolumn{1}{c|}{76.54}   & \multicolumn{1}{c|}{77.65}   & 85.88    \\ \cline{2-7} 
                         & Proposed & 1.0  & 85.31 & \multicolumn{1}{c|}{76.73}   & \multicolumn{1}{c|}{78.62}   & 88.85    \\ \hline
\multirow{2}{*}{ResNet9} &  \cite{OptimMUL2}        & 0    & 76.93 & \multicolumn{1}{c|}{86.10}   & \multicolumn{1}{c|}{48.44}   & 90.60    \\ \cline{2-7} 
                         & Proposed & 1.13 & 85.89 & \multicolumn{1}{c|}{86.38}   & \multicolumn{1}{c|}{61.62}   & 91.24    \\ \hline
\end{tabular}
\caption{Comparison of model accuracy between layer-wise partial-updates induced optimization-based unlearning and its naive counterpart on targeted and retained classes of OrganAMNIST data set.}\label{tabel6}
\vspace{-3mm}
\end{table}

The rationale behind the observed phenomenon is rooted in the approach of naive optimization-based unlearning, where a brief retraining is conducted on the targeted data with the objective of maximizing the loss rather than minimizing it. Consequently, the resulting model not only loses all knowledge of the target class but also has a detrimental impact on the fundamental representations, shared among different classes. This is particularly problematic, as it introduces unintended distortions in the learned features of the retained classes. In contrast, the utilization of layer-wise partial-updates acts as a mitigating factor against this adverse effect. This is achieved by deliberately slowing down the process of modifying the representations of retained classes through partial updates. Partial updates in optimization-based unlearning ensure that erasing knowledge of the targeted class is achieved with minimal impact on the representations of retained classes.

\section{Conclusions}
This paper presented novel solutions to address critical challenges in current machine unlearning methods. One of our proposed methods, partial amnesiac unlearning, integrates layer-wise pruning with amnesiac unlearning, offering a compelling approach to forget specific data from trained models. By subtracting the pruned updates, as opposed to the entire update made during the training, partial amnesiac unlearning minimizes the adverse effects on the model's behavior. Consequently, the model after partial amnesiac unlearning exhibits better performance on the retained data set and requires no brief fine-tuning, unlike in the case of conventional amnesiac unlearning. Additionally, the integration of layer-wise partial-updates into label-flipping and optimization-based unlearning methods minimize the adverse effects on the representation of retained classes. The utilization of layer-wise partial-updates mitigates this adverse effect by slowing down the process of modifying the representations of retained classes through partial updates. The comprehensive experimental assessments across diverse data sets and neural network architectures have showcased the proficiency of the proposed class of unlearning methods in effectively erasing targeted data from trained models. Importantly, these methods have demonstrated their superiority to preserve model performance on retained data compared to their naive counterparts. In near future, we will explore the layer-wise structured and adaptive partial machine unlearning to erase targeted data from trained models.

\bibliographystyle{unsrtnat}
\bibliography{ref}

\end{document}